%% file: main.tex
\documentclass{article}
\usepackage{iclr2024_conference,times}
\usepackage{xcolor}
\input{math_commands.tex}

\usepackage[colorlinks, linkcolor=blue , anchorcolor=red, citecolor=cyan, breaklinks]{hyperref}
\usepackage{url}
\usepackage{arydshln}
\usepackage{soul}

\usepackage[utf8]{inputenc} 
\usepackage[T1]{fontenc}    
\usepackage{booktabs}       
\usepackage{amsfonts}       
\usepackage{nicefrac}      
\usepackage{microtype}      
\usepackage{xcolor}         
\usepackage{microtype}
\usepackage{graphicx}
\usepackage{tabularx}
\usepackage{subfigure}
\usepackage{booktabs}
\usepackage{hyperref}
\usepackage{color}
\usepackage{xcolor}

\usepackage{amsmath}
\usepackage{amssymb}
\usepackage{mathtools}
\usepackage{amsthm}
\usepackage[capitalize,noabbrev]{cleveref}
\usepackage{algorithm}
\usepackage{algorithmic}

\usepackage[textsize=tiny]{todonotes}
\usepackage{multirow}
\usepackage{float}
\usepackage{stfloats}
\usepackage{times}
\usepackage{epsfig}

\usepackage{wrapfig}
\usepackage{enumitem}

\usepackage{subfiles}

\usepackage{color}
\usepackage{colortbl}
\setlist[itemize]{leftmargin=2em}

\usepackage{wrapfig}

\usepackage{subfigure} 
\usepackage{amssymb}
\usepackage{amsbsy}
\usepackage{multirow}
 
\usepackage{wrapfig}
\usepackage{bbding}
\usepackage{color}
\usepackage{fancyhdr}

\renewcommand{\algorithmicrequire}{\textbf{Input:}}
\renewcommand{\algorithmicensure}{\textbf{Output:}}
\newcommand{\model}{MiniRAG}
\newcommand{\dataset}{LiHuaWorld}
\newcommand{\cI}{\mathcal{I}}

\newcommand{\cR}{\mathcal{R}}

\newcommand{\ie}{\textit{i}.\textit{e}.}
\newcommand{\eg}{\textit{e}.\textit{g}.}

\usepackage{wrapfig}

\title{\includegraphics[scale=0.03]{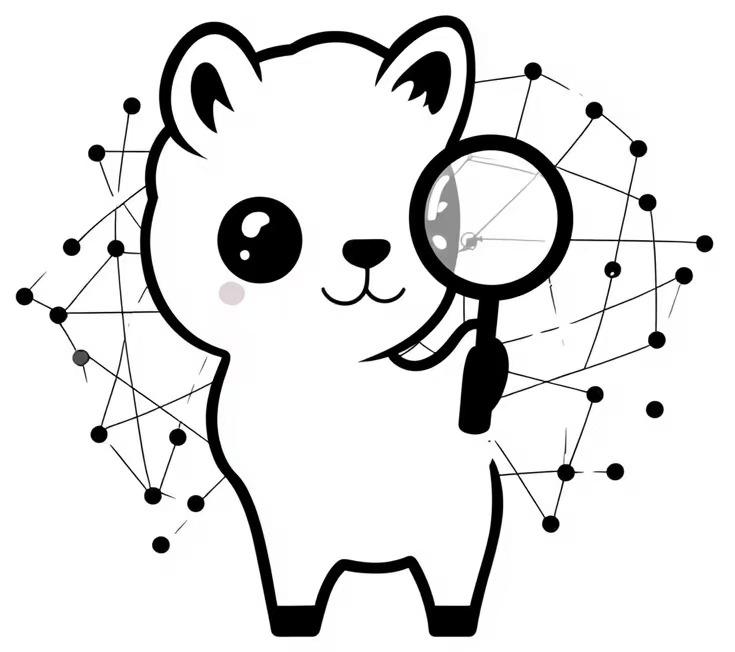}MiniRAG: Towards Extremely Simple \\ Retrieval-Augmented Generation~~~~~~}

\iclrfinalcopy

\author{Tianyu Fan, Jingyuan Wang, Xubin Ren, Chao Huang\thanks{Chao Huang is the corresponding author}\\
University of Hong Kong\\
{\tt $\{$tianyufan0504,jameswangjingyuan,xubinrencs,chaohuang75$\}$@gmail.com}; 
}

\begin{document}

\maketitle

\begin{abstract}
The growing demand for efficient and lightweight Retrieval-Augmented Generation (RAG) systems has highlighted significant challenges when deploying Small Language Models (SLMs) in existing RAG frameworks. Current approaches face severe performance degradation due to SLMs' limited semantic understanding and text processing capabilities, creating barriers for widespread adoption in resource-constrained scenarios. To address these fundamental limitations, we present \model\, a novel RAG system designed for extreme simplicity and efficiency. \model\ introduces two key technical innovations: (1) a semantic-aware heterogeneous graph indexing mechanism that combines text chunks and named entities in a unified structure, reducing reliance on complex semantic understanding, and (2) a lightweight topology-enhanced retrieval approach that leverages graph structures for efficient knowledge discovery without requiring advanced language capabilities. Our extensive experiments demonstrate that \model\ achieves comparable performance to LLM-based methods even when using SLMs while requiring only 25\% of the storage space. Additionally, we contribute a comprehensive benchmark dataset for evaluating lightweight RAG systems under realistic on-device scenarios with complex queries. We fully open-source our implementation and datasets at: \textcolor{blue}{\url{https://github.com/HKUDS/MiniRAG}}.
\end{abstract}

\input{intro}
\input{solution}

\input{evaluation}
\input{relatedworks}
\input{conclusion}
\clearpage

\bibliography{reference}
\bibliographystyle{iclr2024_conference}

\appendix

\clearpage
\input{appendix.tex}

\clearpage

\end{document}

%% file: math_commands.tex
\usepackage{amsmath,amsfonts,bm}

\def\eqref#1{equation~\ref{#1}}

\def\1{\bm{1}}

\DeclareMathAlphabet{\mathsfit}{\encodingdefault}{\sfdefault}{m}{sl}
\SetMathAlphabet{\mathsfit}{bold}{\encodingdefault}{\sfdefault}{bx}{n}

%% file: intro.tex
\section{Introduction}
\label{sec:intro}

Recent advances in Retrieval-Augmented Generation (RAG) have revolutionized how language models access and utilize external knowledge, demonstrating impressive capabilities across diverse applications from question answering to document synthesis~\citep{fan2024survey}. While these systems achieve remarkable performance through sophisticated retrieval mechanisms and powerful language models, they predominantly rely on Large Language Models (LLMs) throughout their pipeline - from index construction and knowledge retrieval to final response generation~\citep{gao2023retrieval}. This extensive dependence on LLMs introduces substantial computational overhead and resource requirements, creating significant barriers for deployment in resource-constrained scenarios such as edge devices, privacy-sensitive applications, and real-time processing systems~\citep{liu2024mobilellm}. Despite growing demand for efficient and lightweight language model applications, current RAG frameworks offer limited solutions for maintaining robust performance within these practical constraints, highlighting a critical gap between theoretical capabilities and real-world deployment needs.

The limitations of existing RAG systems become particularly apparent when attempting to utilize Small Language Models (SLMs) for resource-efficient deployment. While these compact models offer significant advantages in terms of computational efficiency and deployment flexibility, they face fundamental challenges in key RAG operations - from semantic understanding to effective information retrieval. Current RAG architectures (\eg, LightRAG~\cite{guo2024lightrag} and GraphRAG~\cite{edge2024local}), originally designed to leverage LLMs' sophisticated capabilities, fail to accommodate the inherent constraints of SLMs across multiple critical functions: sophisticated query interpretation, multi-step reasoning, semantic matching between queries and documents, and nuanced information synthesis. This architectural mismatch manifests in two significant ways: either severe performance degradation where accuracy drops, or complete system failure where certain advanced RAG frameworks become entirely inoperable when transitioning from LLMs to SLMs.

To address these fundamental challenges, we propose \model, a novel RAG system that reimagines the information retrieval and generation pipeline with a focus on extreme simplicity and computational efficiency. Our design is motivated by three fundamental observations about Small Language Models (SLMs): (1) while they struggle with sophisticated semantic understanding, they excel at pattern matching and localized text processing; (2) explicit structural information can effectively compensate for limited semantic capabilities; and (3) decomposing complex RAG operations into simpler, well-defined steps can maintain system robustness without requiring advanced reasoning capabilities. These insights lead us to prioritize structural knowledge representation over semantic complexity, marking a significant departure from traditional LLM-centric RAG architectures.

Our \model\ introduces two key technical innovations that leverage these insights: (1) a semantic-aware heterogeneous graph indexing mechanism that systematically combines text chunks and named entities in a unified structure, reducing reliance on complex semantic understanding, and (2) a lightweight topology-enhanced retrieval approach that utilizes graph structures and heuristic search patterns for efficient knowledge discovery. Through careful design choices and architectural optimization, these components work synergistically to enable robust RAG functionality even with limited model capabilities, fundamentally reimagining how RAG systems can operate within the constraints of SLMs while leveraging their strengths.

Through extensive experimentation across datasets and Small Language Models, we demonstrate \model's exceptional performance: compared to existing lightweight RAG systems, \model\ achieves 1.3-2.5× higher effectiveness while using only 25\% of the storage space. When transitioning from LLMs to SLMs, our system maintains remarkable robustness, with accuracy reduction ranging from merely 0.8\% to 20\% across different scenarios. Most notably, \model\ consistently achieves state-of-the-art performance across all evaluation settings, including tests on two comprehensive datasets with four different SLMs, while maintaining a lightweight footprint suitable for resource-constrained environments such as edge devices and privacy-sensitive applications. To facilitate further research in this direction, we also introduce LiHuaWorld, a comprehensive benchmark dataset specifically designed for evaluating lightweight RAG systems under realistic on-device scenarios such as personal communication and local document retrieval.

%% file: solution.tex
\section{The \model\ Framework}
\label{sec:solution}
\begin{figure*}[t]
    \begin{center}
    \includegraphics[width=1\textwidth]{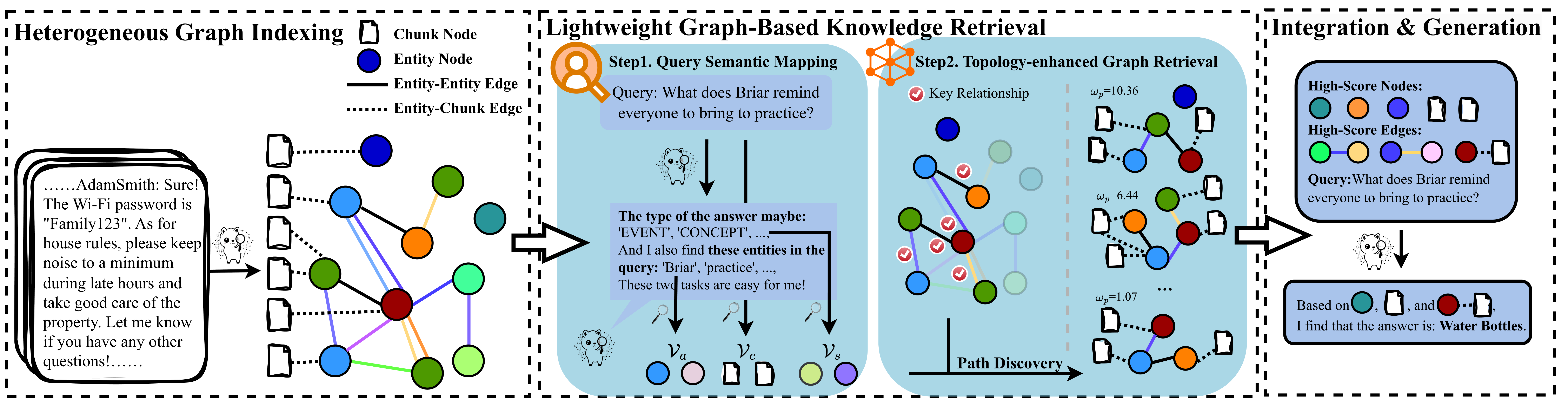}
    \end{center}

    \caption{The \model\ employs a streamlined workflow built on the key components: heterogeneous graph indexing and lightweight graph-based knowledge retrieval. This architecture addresses the unique challenges faced by on-device RAG systems, optimizing for both efficiency and effectiveness.
    }
    \label{Fig:model}
\end{figure*}

In this section, we present the detailed architecture of our proposed \model\ framework. As illustrated in Fig.\ref{Fig:model}, \model\ consists of two key components: (1) heterogeneous graph indexing (Sec.\ref{sec:indexing}), which creates a semantic-aware knowledge representation, and (2) lightweight graph-based knowledge retrieval (Sec.\ref{sec:retrieval}), which enables efficient and accurate information retrieval.

\subsection{Heterogeneous Graph Indexing with Small Language Models}\label{sec:indexing}
In resource-constrained RAG systems, Small Language Models (SLMs) introduce significant operational constraints that impact their effectiveness. These limitations primarily manifest in two critical areas: i) reduced capability to extract and understand complex entity relationships and subtle contextual connections from raw text corpus, and ii) diminished capacity to effectively summarize large volumes of text and process retrieved information containing noise and irrelevant content.

\begin{figure*}[!ht]
    \begin{center}
    \includegraphics[width=0.95\textwidth]{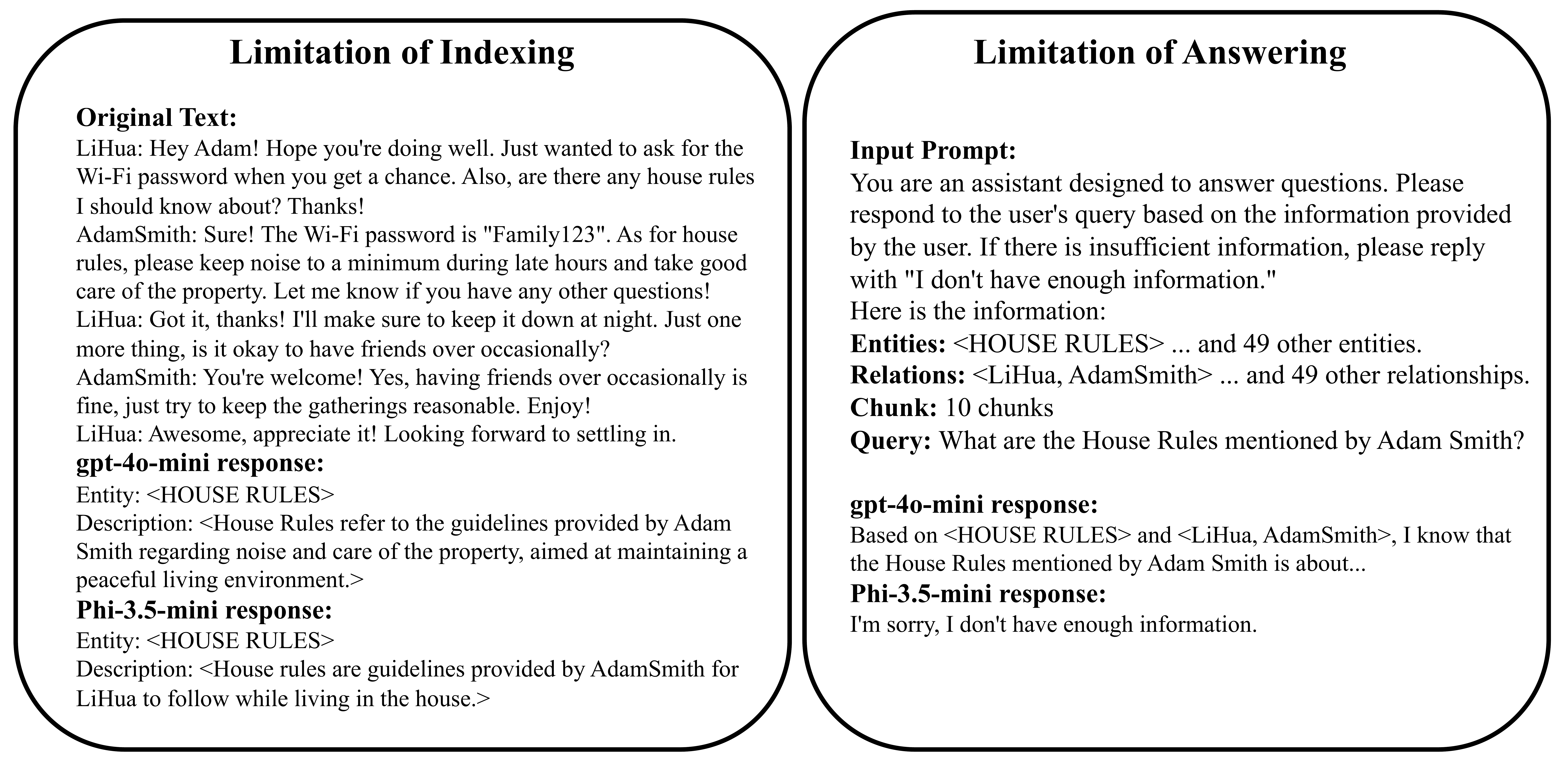}
    \end{center}
    \vspace{-1em}
    \caption{Compared to Large Language Models (LLMs), Small Language Models (SLMs) show significant limitations in both indexing and answering phases. \textbf{Left:} SLMs generate notably lower-quality descriptions than LLMs. \textbf{Right:} When processing identical inputs, SLMs struggle to locate relevant information in large contexts, while LLMs perform this task effectively.}
    \label{houserule}
 \end{figure*}

As shown in Fig.\ref{houserule}, comparing SLM (Phi-3.5-mini~\citep{abdin2024phi}) with LLM (gpt-4o-mini~\citep{openai2023gpt4}) reveals these limitations in practice. While both models identify the "HOUSE RULES" entity, the SLM's description lacks specific details and fails to capture the rules and purposes present in the original text (Limitation 1). Furthermore, during the answering phase, SLMs struggle to locate relevant information within extensive contexts, often becoming distracted by irrelevant content - a challenge not faced by LLMs (Limitation 2).

To address the challenges in resource-constrained RAG systems effectively, our \model\ aims to:
\begin{itemize}[leftmargin=*]

\item The indexing mechanism should extract the key relationships and contextual connections within the data, despite the small models' reduced entity understanding and summarization capacity.

\item The indexing approach should condense the retrieved content to its most query-relevant elements, thereby minimizing potential distractions or misleading information that could impairs the small model's capabilities in both summarization and effective denoising of the retrieved content.

\end{itemize}

To achieve the above goals, we propose a data indexing mechanism that generates a \textbf{Semantic-Aware Heterogeneous Graph}. This graph structure systematically incorporates both text chunks and named entities extracted from the raw text, creating a rich semantic network that facilitates precise information retrieval. In the constructed heterogeneous graph, the nodes comprise two primary types:
\begin{itemize}[leftmargin=*]
\item \textbf{Text Chunk Node $\mathcal{V}_{c}$}: Coherent segments of the original text that preserve contextual integrity.
\item \textbf{Entity Node $\mathcal{V}_{e}$}: The key semantic elements extracted from chunks include events, locations, temporal references, and domain-specific concepts that serve to anchor semantic understanding.
\end{itemize}
This dual-node design enables data chunks to directly participate in the retrieval stage, ensuring identification of the most contextually relevant content. This approach effectively mitigates information distortion that could arise from the limited summarization capabilities of the small language model. Within the heterogeneous graph, the connecting edges between nodes fall into two fundamental types:
\begin{itemize}[leftmargin=*]
\item \textbf{Entity-Entity Connections $\mathcal{E}_{\alpha}$}: Links between named entities that capture semantic relationships, hierarchical structures, and temporal or spatial dependencies.
\item \textbf{Entity-Chunk Connections $\mathcal{E}_{\beta}$}: Bridges between named entities and their corresponding contexts from which the entities are extracted, preserving contextual relevance and semantic coherence.
\end{itemize}
These connections and inter-dependencies are established through language models' semantic understanding capabilities. For example, when \textit{indexing a document that plans a trip to the 2024 Paris Olympics}, the model might establish entity-entity connections between venue locations (Stade de France), event schedules (swimming finals), transportation options (Metro Line 13), and nearby attractions (Eiffel Tower), while creating entity-chunk connections linking these entities to relevant text segments discussing ticket availability, local accommodation reviews, and optimal travel routes.

To further facilitate the relational semantic understanding, we enhance each edge in the knowledge graph with semantic descriptions generated by language models. These descriptions provide explicit relationship context between connected nodes. Specifically, for each edge $e_{\beta} \in \mathcal{E}_{\beta}$ that connects an entity to its corresponding chunk, we employ the language model to generate a description $d_{e_{\beta}}$ of the entity as supplementary information for this edge. This description provides detailed content about the entity and reflects the semantic relationship between the extracted entity and the relevant chunk. With the textual description of the entity-chunk edge, it becomes a text-attributed edge $(e_{\beta}, d_{e_{\beta}}) \in \mathcal{E}_{\beta}$. In summary, the indexing process within our \model\ framework yields a heterogeneous graph $\mathcal{G}$ that encompasses both entity and chunk nodes with semantic-aware connections as follows:
\begin{align}
    \mathcal{D} = \mathcal{G} = (\{\mathcal{V}_{c}, \mathcal{V}_{e}\}, \{ \mathcal{E}_{\alpha}, (e_{\beta}, d_{e_{\beta}}) \in \mathcal{E}_{\beta}\})
\end{align}

\subsection{Lightweight Graph-Based Knowledge Retrieval}
\label{sec:retrieval}
In on-device Retrieval Augmented Generation (RAG) systems, the limitations of device computational capabilities and data privacy restrict the use of powerful models, such as large language models and advanced text embedding models, necessitating reliance on smaller alternatives. Consequently, currently used pipelines heavily rely on LLMs for a comprehensive understanding of text semantics when computing embedding similarity for retrieval, facing significant challenges. These smaller models often struggle to capture the precise semantic nuances within lengthy texts, complicating accurate matching. To tackle these challenges, it is essential to: i) \textbf{Reduce the complexity of input content} for generation, ensuring that semantic information is clear and concise; ii) \textbf{Shorten the length of input content} for smaller language models, facilitating improved comprehension and retrieval accuracy. Additionally, employing effective graph indexing structures can help mitigate performance deficiencies in semantic matching, thereby enhancing the overall retrieval process.

In \model, we propose a \textbf{Graph-based Knowledge Retrieval} mechanism that effectively leverages the semantic-aware heterogeneous graph $\mathcal{G}$ constructed during the indexing phase, in conjunction with lightweight text embeddings, to achieve efficient knowledge retrieval. By employing a graph-based search design, we aim to ease the burden on precise semantic matching with large language models. This approach facilitates the acquisition of rich and accurate textual content at a low computational cost, thereby enhancing the ability of language models to generate precise responses.

\subsubsection{Query Semantic Mapping}
In the retrieval phase, the primary goal for a user-input query \( q \) is to identify elements related to the query (e.g., named entities, text chunks) from the constructed index data, thereby aiding the model in generating accurate responses. To achieve this, it is essential to first parse the query and align it with the index data. Some prior RAG methods utilize LLMs to expand or decompose the query into fine-grained queries~\citep{chan2024rq, edge2024local, guo2024lightrag}, enhancing the match between the query and the index data. However, this process relies on LLMs to extract high-quality abstract information from the query, which poses challenges for smaller language models. Therefore, in the retrieval process of \model, we leverage entity extraction—a relatively simple and effective task for small language models—to facilitate the decomposition and mapping of the query \( q \) to our graph-based indexing data (i.e., the semantic-aware heterogeneous graph \( \mathcal{G} \)).

For a given \( q \), our approach begins with a two-stage entity processing pipeline. First, we employ a small language model to extract relevant entities
\( \mathcal{V}_{q} \) from \( q \) while simultaneously predicting their potential types (\eg, event, location, person) that may directly contribute to the query's answer. Following this, we leverage a lightweight sentence embedding model to evaluate semantic similarities across all entity nodes \( \mathcal{V}_{e} \) in the constructed graph $\mathcal{G} = \{\mathcal{V}_{c}, \mathcal{V}_{e}\}$, examining various text corpora (\ie, entity names, chunk content) to enable effective node retrieval and grounding.

\noindent \textbf{Query-Driven Reasoning Path Discovery}. Within our semantic-aware heterogeneous graph $\mathcal{G} = \{\mathcal{V}_{c}, \mathcal{V}_{e}\}$, \model\ constructs reasoning paths through an intelligent query-guided mechanism. For any input query $q$, the model identifies relevant text chunks by jointly considering two key aspects: (1) semantic relevance between query and entity nodes, and (2) structural coherence among entity-entity and entity-chunk relationships. This dual-objective optimization framework simultaneously maximizes $q$-$\mathcal{V}{e}$ semantic alignment and preserves ($\mathcal{V}_{e}$-$\mathcal{V}_{e}$), ($\mathcal{V}_{e}$-$\mathcal{V}_{c}$) relational dependencies, effectively capturing complex reasoning chains within the heterogeneous knowledge graph. The systematic query-relevant reasoning path discovery procedure consists of the following key steps:

\begin{itemize}[leftmargin=*]
    \item \textbf{Initial Entity Identification} ($\hat{\mathcal{V}}_s$): We locate high-confidence starting points by matching query entities with graph nodes, establishing reliable entry points for path exploration.
    
    \item \textbf{Answer-Aware Entity Selection} ($\hat{\mathcal{V}}_a$): Leveraging predicted answer types, we identify candidate answer nodes from the starting set, enabling type-guided reasoning.

    \item \textbf{Context-Rich Path Formation} ($\hat{\mathcal{V}}_c$): We enrich reasoning paths by incorporating relevant text chunks, creating comprehensive evidence chains that connect query entities to potential answers.
\end{itemize}

This lightweight framework maintains high efficiency while ensuring semantic accuracy, making it particularly suitable for edge computing scenarios. The subsequent section details our search algorithm for further refining these reasoning paths through importance-based ranking.

\subsubsection{Topology-Enhanced Graph Retrieval}
To address the fundamental limitations of small language models-based methods in knowledge retrieval, we propose a topology-aware retrieval approach that effectively combines semantic and structural information from heterogeneous knowledge graphs. Approaches relying on small language models with limited semantic understanding capabilities, often introduce substantial noise into the retrieval process due to their constrained ability to capture nuanced meanings, contextual variations, and complex entity relationships in real-world knowledge graphs. Our method overcomes these inherent challenges through a carefully designed two-stage process that synergistically leverages both embedding-based similarities and topological structure of the knowledge graph.

The process begins with embedding-based similarity search to identify seed entities ($\hat{\mathcal{V}}_s$, $\hat{\mathcal{V}}_a$) through semantic matching, followed by a topology-enhanced discovery phase that leverages the heterogeneous graph structure $\mathcal{G}$ to discover relevant reasoning paths. By integrating entity-specific relevance scores, structural importance metrics, and path connectivity patterns, our approach achieves superior precision in knowledge retrieval while maintaining computational efficiency, ultimately enabling more accurate and interpretable reasoning paths for enhanced generation tasks.

\begin{itemize}[leftmargin=*]
    \item \textbf{Key Relationship Identification}: We first identify high-quality entity-entity connections within graph $\mathcal{G}$ that are relevant to query $q$ through node-edge interactions. In the entity-entity connections $\mathcal{E}_\alpha$, we define an edge as highly relevant if it connects a starting node $\hat{v}_s \in \hat{\mathcal{V}}_s$ to an answer node $\hat{v}_a \in \hat{\mathcal{V}}_a$ along their shortest path. For efficient extraction, we focus on edges proximate to starting or answer nodes. The relevance scoring function $\omega_e (\cdot)$ for each edge $e_\alpha \in \mathcal{E}_{\alpha}$ is formally defineds:
    \begin{align}
        \omega_e (e) = \sum_{\hat{v}_s \in \hat{\mathcal{V}}_s} \text{count}(\hat{v}_s , \hat{\mathcal{G}}_{e,k}) + \sum_{\hat{v}_a \in \hat{\mathcal{V}}_a} \text{count}(\hat{v}_a, \hat{\mathcal{G}}_{e,k}),
    \end{align}
    where $\hat{\mathcal{G}}_{e,k}$ denotes the $k$-hop subgraph centered at edge $e$, encompassing all nodes and edges reachable within $k$ steps from either endpoint. Based on the computed relevance scores $\omega_e$, we construct the key relationships set $\hat{\mathcal{E}}_\alpha$ by selecting the top-ranked edges.

    \item \textbf{Query-Guided Path Discovery}: To systematically discover logically relevant information within our knowledge graph structure, we identify and extract significant paths that serve as meaningful reasoning chains. A reasoning path starts from a carefully selected seed node and progressively advances toward potential answer nodes while maximizing the incorporation of previously extracted key relationships. For each candidate starting node $\hat{v}_s$ in our graph, we comprehensively define the potential reasoning path set $\mathcal{P}{\hat{v}_s}$ as the collection of all possible acyclic paths of length $n$ originating from $\hat{v}_s$. For each identified query entity $v_q \in \mathcal{V}_q$, we systematically evaluate these potential paths using a sophisticated entity-conditioned score function $\omega_p(\cdot)$ that quantifies both the overall path importance and query relevance through multiple dimensions:
    \begin{align}
        \omega_p (p \mid v_q) &= \omega_v (\hat{v}_s \mid v_q) \cdot (1+\sum_{v \in (p \land \hat{\mathcal{V}}_{a})} \text{count}(v , p) + \sum_{e \in( p \land \hat{\mathcal{E}}_\alpha)} \omega_e (e)).
    \end{align}
    The scoring components in our path discovery framework are defined as follows: $\omega_v (\hat{v}_s \mid v_q)$ measures the semantic similarity between starting node $\hat{v}_s$ and query entity $v_q$ using cosine similarity of their respective embeddings in the vector space, while $\text{count}(v, p)$ serves as a binary indicator function that returns 1 if node $v$ appears in path $p$ and 0 otherwise. For each query entity and starting node pair in the graph, we systematically rank all potential paths according to their computed importance scores and construct the final comprehensive set of reasoning paths $\mathcal{P}_{q}$ by carefully selecting the top-$k$ highest-scoring paths from each ranking list for subsequent steps.

    \item \textbf{Retrieval of Query-Relevant Text Chunks}: Building upon our indexing structure from Section~\ref{sec:indexing}, each entity node maintains connections with its source text chunk through entity-chunk inter-dependencies. These text chunks exist as nodes in our indexing graph, connected via text-attributed edges $(e_\beta, d_{e_\beta}) \in \mathcal{E}_\beta$. By traversing these connections, we collect all chunk nodes $\mathcal{V}_c^q$ that are linked to entity nodes present in any reasoning path $p \in \mathcal{P}_{q}$. \textbf{Step 1: Candidate Filtering.} We first systematically filter the candidates to focus on the intersection $\hat{\mathcal{V}}_c \land \mathcal{V}_c^q$ to ensure coverage of relevant information. \textbf{Step 2: Similarity Computation.} For each candidate chunk in the intersection, we carefully calculate the semantic similarity between the input query and the concatenated content, which combines both the chunk text and its associated edge descriptions. \textbf{Step 3: Ranking and Selection.} Finally, we rank these filtered chunks according to their computed relevance scores and select the top-scoring candidates to form the final optimized set $\hat{\mathcal{V}}_c^q$ for subsequent reasoning.

    \item \textbf{Integration for Augmented Generation}: Through our proposed topology-enhanced graph retrieval mechanism and multi-stage filtering process, we efficiently obtain two key components of query-relevant graph knowledge: i) Essential relationships $\hat{\mathcal{E}}_\alpha$ connecting important entities within the knowledge graph, which capture the semantic dependencies and structural patterns; ii) Optimal text chunks $\hat{\mathcal{V}}_c^q$ containing critical contextual information and supporting evidence necessary for accurate answer generation. By systematically integrating these retrieved components with the previously grounded answer nodes $\hat{\mathcal{V}}_a$ through our designed fusion strategy, we construct the comprehensive and well-structured input representation for the final augmented generation process.
    
\end{itemize}

%% file: evaluation.tex
\section{Evaluation}

Through the novel design of \model, we enable efficient on-device RAG systems without relying on large models, preserving data privacy while maintaining robust performance. Our evaluation addresses three key research questions (RQs): $\bullet$ \textbf{RQ1: Comparative Performance.} How does \model\ perform against state-of-the-art alternatives in terms of retrieval accuracy and efficiency? $\bullet$ \textbf{RQ2: Component Analysis.} What is the contribution of key components to \model's overall effectiveness? $\bullet$ \textbf{RQ3: Case Studies.} How effectively does \model\ handle complex, multi-step reasoning tasks with small language models, as demonstrated through practical case studies?

\subsection{Experimental Settings}

\textbf{Datasets.} The evaluation of on-device RAG requires careful consideration of their unique context and practical use cases. While traditional server-side RAG systems are designed to process extensive documents such as academic papers, technical reports, and comprehensive web content, on-device RAG applications serve fundamentally different purposes aligned with users' daily device interactions.

Our dataset selection reflects these requirements, focusing on common on-device scenarios including \textbf{Instant Messaging} (real-time retrieval from chat histories and personal communications, and emails), \textbf{Personal Content} (user-created notes, memos, and calendar entries), and \textbf{Local Short Documents} (lightweight text files). This composition aligns with the core strengths of on-device RAG systems - privacy preservation, real-time processing, and efficient personal content management. By focusing on these everyday user interactions rather than complex document processing, our evaluation framework provides a realistic assessment of on-device RAG capabilities in their intended use cases.

Our evaluation employs two datasets (detailed in Appendix Section) that capture essential aspects of real-world, on-device RAG scenarios. The key characteristics of these datasets are as follows:

\noindent $\bullet$ \textbf{Synthetic Personal Communication Data}: To comprehensively capture real-world personal communications, we leverage GPT-4 to generate a year-long dataset that authentically mirrors the full spectrum of daily life interactions. This extensive dataset encompasses diverse aspects of modern living, including daily necessities (\eg, food, clothing, housing, transportation), social activities and entertainment, work and study-related discussions, personal schedule planning, and shopping decisions. The conversations reflect natural communication patterns across various contexts - from casual chats and task coordination to information sharing and decision making. By utilizing GPT-4's advanced generation capabilities with effective prompting mechanism, we ensure the dataset maintains realistic temporal coherence and contextual relationships while preserving privacy, making it ideal for evaluating both ``Instant Messaging'' and ``Personal Content'' use cases.

\noindent $\bullet$ \textbf{Short Documents}: We utilize a multi-hop RAG dataset based on contemporary news articles, specifically designed to evaluate the system's capability in navigating and retrieving information across multiple short documents. This dataset mirrors the real-world scenarios, where users frequently need to retrieve relevant information from various locally stored files. This setup enables comprehensive assessment of both retrieval efficiency and accuracy when handling cross-document information access - for the ``Local Short Documents'' use case in on-device applications.

\textbf{Evaluation Protocols and Metrics.} We employ two key metrics to assess the quality and reliability of responses generated by various RAG methods. $\bullet$ \textbf{Accuracy ($acc$)}: Measures the consistency between the RAG system's response and the expected answer. For instance, given the query ``What does Briar remind everyone to bring to practice?'' with the expected answer ``bottle'', semantically equivalent responses like ``water bottle'' are considered correct. $\bullet$ \textbf{Error Rate ($err$)}: Captures instances where the RAG system provides incorrect information without recognizing its mistake. For example, if the system confidently responds with ``yoga mat'' to the above query, it would count toward the error rate.

\textbf{Implementation Details.} We configure our experimental setup following established practices from prior work~\citep{guo2024lightrag}. For text processing, we set the chunk size to 1200 tokens with an overlap of 100 tokens, and utilize \text{nano vector base} for vector storage. In our \model\ implementation, we configure the top-k retrieval to 5 documents and set the maximum token limit to 6000 tokens.

For the model selection, we employ different efficient configurations for large and small language models. In the advanced LLM setting, we use gpt-4o-mini~\citep{openai2023gpt4} as the language model and text-embedding-3-small as the specialized embedding model. For the lightweight SLM setting, we utilize optimized all-MiniLM-L6-v2~\citep{reimers-2019-sentence-bert} as the embedding model, paired with various small language models including Phi-3.5-mini-instruct~\citep{abdin2024phi}, GLM-Edge-1.5B-Chat, Qwen2.5-3B-Instruct~\citep{qwen2.5}, and MiniCPM3-4B~\citep{hu2024minicpm}.

\textbf{Baselines.} We compare our \model\ against several representative RAG systems:

\begin{itemize}[leftmargin=*]
    
\item{\textbf{NaiveRAG}}~\citep{mao2020generation} serves as the standard RAG baseline, employing text embedding-based retrieval. It segments documents into chunks stored in a vector database and performs retrieval through direct similarity matching between query and chunk embeddings.

\item{\textbf{GraphRAG}}~\citep{edge2024localglobalgraphrag} leverages graph-based indexing through language models and the Leiden algorithm for entity clustering. It enhances retrieval by generating community reports and combining local-global information access through a unified retrieval mechanism.

\item{\textbf{LightRAG}}~\citep{guo2024lightrag} implements a dual-level retrieval architecture with knowledge graphs. It enhances query understanding by decomposing queries into hierarchical components (low-level details and high-level concepts), enabling more precise document retrieval.

\end{itemize}

\input{table_mainexp}

\subsection{Performance Analysis (RQ1)}

$\bullet$ \textbf{Performance Degradation in Existing RAG Systems with SLMs
}. Current RAG systems face critical challenges when operating with Small Language Models (SLMs), revealing fundamental vulnerabilities in their architectures. Advanced LLM-based RAG methods exhibit severe performance degradation, with LightRAG's accuracy plummeting from 56.90\% to 35.42\% during LLM to SLM transition, while GraphRAG experiences complete system failure due to its inability to generate high-quality content. While basic retrieval systems like NaiveRAG show resilience, they suffer from significant limitations, being restricted to basic functionality and lacking advanced reasoning capabilities. This performance analysis highlights a critical challenge: existing advanced systems' over-reliance on sophisticated language capabilities leads to fundamental operational failures when using simpler models, creating a significant barrier to widespread adoption in resource-constrained environments, where high-end language models may not be available or practical to deploy.

$\bullet$ \textbf{\model's Unique Advantages}. These innovations enable MiniRAG to maintain strong performance even with simpler language models, making it particularly suitable for resource-constrained environments while preserving the core functionalities of RAG systems. 

\textbf{i) Semantic-Aware Graph Indexing for Reduced Model Dependency}. \model\ fundamentally reimagines the indexing process through a dual-node heterogeneous graph structure. Instead of relying on powerful text generation capabilities, the system focuses on basic entity extraction and heterogeneous relationship mapping. This design combines text chunk nodes for preserving raw contextual information with entity nodes for capturing key semantic elements, creating a robust knowledge representation that remains effective even with limited language model capabilities.

\textbf{ii) Topology-Enhanced Retrieval for Balanced Performance}. \model\ employs a lightweight graph-based retrieval mechanism that balances multiple information signals through a systematic process. Beginning with query-driven path discovery, the system integrates embedding-based matching with structural graph patterns and entity-specific relevance scores.
Through topology-aware search and optimized efficiency, it achieves robust retrieval quality without requiring advanced language understanding, making it particularly effective for on-device deployment.
\begin{wrapfigure}[9]{r}{0.35\textwidth}
\vspace{-1.5em}
    \begin{centering}
\includegraphics[width=1\linewidth]{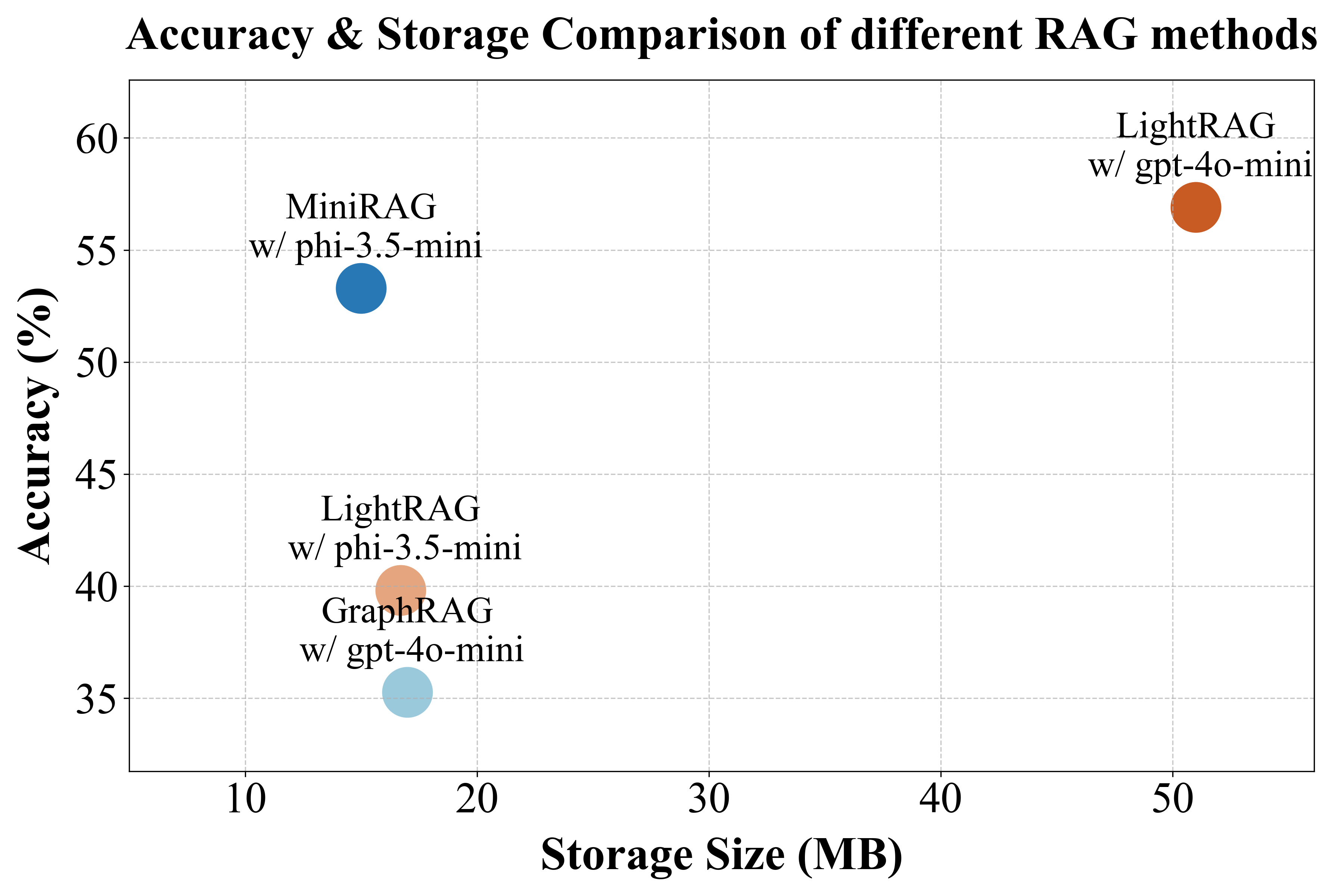}
      \vspace{-2em}
      \caption{Accuracy vs. Storage Efficiency: Comparative analysis of three RAG systems - MiniRAG, LightRAG, and GraphRAG.}
      \label{fig:compare}
    \end{centering}
\end{wrapfigure}
These innovations enable \model\ to maintain strong performance with simpler language models, making it ideal for resource-constrained environments while preserving core RAG functionalities.

$\bullet$ \textbf{Storage Efficiency While Maintaining Performance}. \model\ demonstrates exceptional storage efficiency while preserving high accuracy levels. Empirical evaluations show that \model\ achieves competitive accuracy while requiring only 25\% of the storage space compared to baselines like LightRAG w/ gpt-4o-mini. This dramatic reduction in storage requirements, coupled with maintained or improved accuracy, represents a significant advancement in efficient RAG system design.

\subsection{Component-wise Analysis of \model\ (RQ2)}
Our ablation study examines the individual contributions of key \model\ components through two primary experimental variations, as documented in Table~\ref{tab_ablation}. The first variation (-$\cI$) replaces \model's heterogeneous graph indexing with a description-based indexing approach that requires comprehensive semantic understanding for generating accurate entity/edge descriptions, similar to methods used in LightRAG and GraphRAG. The second variation (-$\cR_i$) involves selectively deactivating specific modules during graph retrieval. This systematic evaluation framework provides detailed insights into how each component contributes to \model's overall performance.

Our experimental results reveal crucial insights into \model's architectural effectiveness. $\bullet$ \textbf{Validating SLM Limitations}. One key finding emerges when replacing our streamlined indexing method with text semantic-driven indexing techniques (-$\cI$), resulting in substantial performance degradation. This outcome strongly validates our initial hypothesis about Small Language Models (SLMs) and their inherent limitations - specifically their constraints in comprehensive semantic understanding, which impacts both the generation of complex knowledge graphs with entity relationships and the creation of corresponding comprehensive text descriptions. $\bullet$ \textbf{Effectiveness of Query-guided Reasoning Path Discovery}. The experiments further demonstrate the critical nature of structural components: the removal of either edge information (-$\cR_{edge}$) or chunk nodes (-$\cR_{chunk}$) significantly impacts system performance. These elements serve dual purposes: they facilitate effective query matching through query-guided reasoning path discovery while simultaneously compensating for the inherent limitations of SLMs during the data indexing phase.

\input{table_ablation}

\subsection{Case Study Analysis (RQ3)}
In this section, we demonstrate \model's practical advantages through a case study with LightRAG, focusing on a complex restaurant identification scenario. This study illustrates how our query-guided reasoning approach, combined with heterogeneous graph indexing, effectively handles multi-constraint queries while overcoming the inherent limitations of small language models.

$\bullet$ \textbf{Challenge: Complex Query Resolution in Restaurant Identification}. We conducted a comparative case study between \model\ and LightRAG using a complex query scenario: \textit{"What is the name of the Italian restaurant where \textit{Wolfgang} and \textit{Li Hua} are having dinner to celebrate Wolfgang's promotion?"} This query presents multiple challenges, requiring the system to identify specific Italian restaurants from various mentions in online chat data while correlating them with the context of a promotion celebration. LightRAG, despite its capabilities, struggled with this task due to the limitations of its underlying small language model (phi-3.5-mini-instruct). The SLM's constraints in extracting appropriate high-level information, combined with noise in the graph-based index, led to ineffective direct embedding matching and ultimately hindered accurate answer retrieval.

$\bullet$ \textbf{\model's Effective Query-Guided Knowledge Discovery}. MiniRAG successfully resolved this challenge through its query-guided reasoning path discovery mechanism, which enables precise and contextually relevant knowledge retrieval. By leveraging its heterogeneous graph indexing structure, \model\ effectively constructs query-relevant knowledge paths, starting with answer type prediction ("Social Interaction" or "Location") and proceeding through targeted entity matching. This structured reasoning approach, combined with strategic decomposition of query elements (focusing on "Italian place" and "restaurant" contexts), allows \model\ to navigate the knowledge space efficiently. The synergy between query-guided reasoning and heterogeneous graph indexing enabled \model\ to effectively filter through multiple Italian establishments, ultimately identifying \textit{"Venedia Grancaffe"} as the venue specificaslly connected to the promotion celebration context.

\input{table_case}

%% file: table_mainexp.tex
\begin{table}[t]
\centering
\caption{Performance evaluation using accuracy (acc) and error (err) rates, measured as percentages (\%). Higher accuracy and lower error rates indicate better RAG performance. Results compare various baseline methods against our \model\ across multiple datasets. Bold values indicate best performance, while ``/'' denotes cases where methods failed to generate effective responses.}
\label{tab_mainexp}

\resizebox{\textwidth}{!}{
\begin{tabular}{@{}lcccccccc@{}}
\toprule
\multirow{2}{*}{~~~~~~~~\dataset}    & \multicolumn{2}{c}{\textbf{NaiveRAG}} & \multicolumn{2}{c}{\textbf{GraphRAG}} & \multicolumn{2}{c}{\textbf{LightRAG}} & \multicolumn{2}{c}{\textbf{\model}} \\ 
\cmidrule(lr){2-3} \cmidrule(lr){4-5} \cmidrule(lr){6-7} \cmidrule(lr){8-9}
                      & $acc{\color[rgb]{1,0,0}\uparrow}$ & $err{\color[rgb]{0.4, 0.71, 0.376}\downarrow}$ & $acc{\color[rgb]{1,0,0}\uparrow}$ & $err{\color[rgb]{0.4, 0.71, 0.376}\downarrow}$ & $acc{\color[rgb]{1,0,0}\uparrow}$ & $err{\color[rgb]{0.4, 0.71, 0.376}\downarrow}$ & $acc{\color[rgb]{1,0,0}\uparrow}$ & $err{\color[rgb]{0.4, 0.71, 0.376}\downarrow}$ \\
\midrule
Phi-3.5-mini-instruct   & 41.22\%  &   23.20\%  &   /  &   /  &   39.81\%  &   25.39\%  &   \textbf{53.29\%}  &   23.35\% \\
GLM-Edge-1.5B-Chat      &   42.79\%  &   24.76\%  &   /  & /  &   35.74\%  &   25.86\%  &   \textbf{52.51\%}  &   25.71\% \\

Qwen2.5-3B-Instruct   &    43.73\%  &   24.14\%  &   /  &   /  &   39.18\%  &   28.68\%  &   \textbf{48.75\%}  &   26.02\% \\   
MiniCPM3-4B       &  43.42\%  &   17.08\%  &   / &   /  &   35.42\%  &   21.94\%  &   \textbf{51.25\%}  &   21.79\%   \\   
\midrule
gpt-4o-mini  &    46.55\%  &   19.12\%  &   35.27\%  &  37.77\%  &   \textbf{56.90\%}  &   20.85\%  & 54.08\%  &     19.44\%            \\

\midrule\\

\multirow{2}{*}{~~~~~~~~MultiHop-RAG }    & \multicolumn{2}{c}{\textbf{NaiveRAG}} & \multicolumn{2}{c}{\textbf{GraphRAG}} & \multicolumn{2}{c}{\textbf{LightRAG}} & \multicolumn{2}{c}{\textbf{\model}} \\ 
\cmidrule(lr){2-3} \cmidrule(lr){4-5} \cmidrule(lr){6-7} \cmidrule(lr){8-9}
                     & $acc{\color[rgb]{1,0,0}\uparrow}$ & $err{\color[rgb]{0.4, 0.71, 0.376}\downarrow}$ & $acc{\color[rgb]{1,0,0}\uparrow}$ & $err{\color[rgb]{0.4, 0.71, 0.376}\downarrow}$ & $acc{\color[rgb]{1,0,0}\uparrow}$ & $err{\color[rgb]{0.4, 0.71, 0.376}\downarrow}$ & $acc{\color[rgb]{1,0,0}\uparrow}$ & $err{\color[rgb]{0.4, 0.71, 0.376}\downarrow}$ \\
\midrule
Phi-3.5-mini-instruct      &     42.72\%   &       31.34\% &   /     &   /       &    27.03\%       &    11.78\%  &  \textbf{49.96\%}  &    28.44\%  \\
GLM-Edge-1.5B-Chat     &   44.44\%     &     24.26\%  &   /     &   /        &      /     &   /   &     \textbf{51.41\%} &   23.44\%   \\
Qwen2.5-3B-Instruct   &   39.48\%      &      31.69\%  &   /     &   /      &    21.91\%     &13.73\% &      \textbf{48.55\%}   &  33.10\%   \\   
MiniCPM3-4B       &   39.24\%      &   31.42\%        &   /     &   /      &   19.48\%    &   10.41\%   &   \textbf{47.77\%}   & 26.88\%     \\
\midrule
gpt-4o-mini     &   53.60\%           &    27.19\%   & 60.92\%      &   16.86\%&   64.91\%     &    19.37\%              &  \textbf{68.43\%}    & 19.41\%     \\
\bottomrule
\end{tabular}
}
\end{table}

%% file: table_ablation.tex
\begin{table}[t]
\centering
\caption{Ablation study results comparing accuracy ($acc$, ↑) and error rate ($err$, ↓) (\%) across architectural variants: baseline \model\ versus variants with (i) semantic-driven indexing replacement (-$\cI$), (ii) edge information removal (-$\cR_{edge}$), and (iii) chunk nodes removal (-$\cR_{chunk}$). Results validate SLM limitations and the effectiveness of query-guided reasoning path components.}
\label{tab_ablation}

\resizebox{\textwidth}{!}{
\begin{tabular}{@{}lcccccccc@{}}
\toprule
\multirow{2}{*}{~~~~~~~~\dataset}    & \multicolumn{2}{c}{\textbf{\model}} & \multicolumn{2}{c}{\textbf{-$\cI$}} & \multicolumn{2}{c}{\textbf{-$\cR_{chunk}$}} & \multicolumn{2}{c}{\textbf{-$\cR_{edge}$} }\\ 
\cmidrule(lr){2-3} \cmidrule(lr){4-5} \cmidrule(lr){6-7} \cmidrule(lr){8-9}
                      & $acc{\color[rgb]{1,0,0}\uparrow}$ & $err{\color[rgb]{0.4, 0.71, 0.376}\downarrow}$ & $acc{\color[rgb]{1,0,0}\uparrow}$ & $err{\color[rgb]{0.4, 0.71, 0.376}\downarrow}$ & $acc{\color[rgb]{1,0,0}\uparrow}$ & $err{\color[rgb]{0.4, 0.71, 0.376}\downarrow}$ & $acc{\color[rgb]{1,0,0}\uparrow}$ & $err{\color[rgb]{0.4, 0.71, 0.376}\downarrow}$ \\
\midrule
Phi-3.5-mini-instruct     & 53.29\%  &   23.35\%  &   26.02\%  &   19.12\%  &   48.90\%  &   17.40\%  &   50.47\%  &   15.36\%  \\
GLM-Edge-1.5B-Chat     &   52.51\%  &   25.71\%  &   25.08\%  &   31.50\%  &   46.24\%  &   16.77\%  &   47.81\%  &   20.53\%   \\

Qwen2.5-3B-Instruct           & 48.75\%  &   26.02\%  &   24.14\%  &   15.67\%  &   40.91\%  &   16.14\%  &   48.43\%  &   18.65\%   \\
MiniCPM3-4B               &  51.25\%  &   21.79\%  &   26.18\%  &   15.52\%  &   46.39\%  &   15.83\%  &   48.59\%  &   19.44\%    \\

\bottomrule
\end{tabular}
}
\end{table}

%% file: table_case.tex
\begin{table*}[t]
\centering
\caption{Case study comparing LightRAG and \model\ on a complex restaurant identification query, demonstrating how query-guided reasoning path discovery effectively addresses small language model (SLM) limitations in multi-constraint information retrieval tasks.}
\label{table_case}
\small
\begin{tabularx}{\textwidth}{X X}
    \toprule

      \textbf{Query:} What is the name of the Italian restaurant where Wolfgang and Li Hua are having dinner to celebrate Wolfgang's promotion?\\

      \midrule
      \textbf{Ground-Truth:} Venedia Grancaffe\\
      \midrule

    \textbf{LightRAG Source:}

    Given the query, I decomse it to \textbf{low-level information}: <"Wolfgang"><"LiHua"><"Restaurant name"> and \textbf{high-level information}: <"Italian restaurant"><"Dinner celebration"><"Promotion">. Using these information, I find these elements:

 \textbf{Entities:}<"FRIES"><"COLLABORATION"><"WOLFGANGSCHULZ"><"HAILEY'S BAKERY"><"LiHUA"><"JAM SESSION">...

   \textbf{Relations:}<"ORGANIZATION","PIZZA ORDER"><"LIHUA","WOLFGANGSCHULZ">
    <"FARRAH","PIZZA"><"LUNCH","THANECHAMBERS">...

   \textbf{Texts:}<Time: 20260318\_15:30><Time: 20260508\_08:00><Time: 20260514\_14:00><Time: 20261103\_13:00><Time: 20260326\_16:00>...

    \textbf{LightRAG Response:}

Wolfgang and Li Hua are planning to have dinner at \textbf{a pasta joint} to celebrate Wolfgang's promotion...The specific name of the restaurant\textbf{ wasn't listed }in the provided information...

\\
      \midrule
      \textbf{\model~Source:}

          Given the query, I guess the \textbf{answer type} could be:
        <"SOCIAL INTERACTION"><"LOCATION"> and I found the \textbf{entities from query} are: <"Italian restaurant"><"Wolfgang"><"LiHua"><"dinner"><"promotion">. Using these information, I find these elements:

        \textbf{Entities from query:}
        <"ITALIAN PLACE"><"RESTAURANT"><"WOLFGANG"><"WOLFGANG SCHULZ"><"LIHUA"><"DINNER PARTY">...

        \textbf{Candidate answer entity:}
        ...<"ITALIAN PLACE"><"GRATITUDE EXPRESSION"><"NEW PIZZA PLACE DOWNTOWN">...

        \textbf{Texts:} <Time: 20260908\_21:00><Time: 20260529\_17:00><Time: 20260523\_14:00><Time: 20260430\_17:00><Time: 20260419\_21:00>

      \textbf{\model~Response:}

The name of the Italian restaurant where Wolfgang and Li Hua are having dinner to celebrate Wolfgang's promotion is \textbf{Venedia Grancaffe}. Wolfgang mentioned it in a reminder about the dinner planned for 6 PM. They were excited to celebrate his promotion with a feast, with both looking forward to enjoying some delicious food together....

\\

     \bottomrule
\end{tabularx}
\vspace{-0.2in}
\end{table*}

%% file: relatedworks.tex
\section{Related Works}

\textbf{Small Language Models (SLMs).} The emergence of Small Language Models (SLMs) is driven by the growing need for lightweight, efficient and privacy-preserving AI solutions that can operate on edge devices, addressing the limitations of large language models in computational resources and deployment costs~\citep{liu2024mobilellm,wang2024comprehensivesurveysmalllanguage,qwen2.5,abdin2024phi,hu2024minicpm}. Recent developments have produced notable models such as MiniCPM3-4B~\citep{hu2024minicpm}, phi-3.5-mini~\citep{abdin2024phi}, Llama-3.2-3B~\citep{grattafiori2024llama3herdmodels}, Qwen2.5-1.5B~\citep{qwen2.5}, gemma-2-2b~\citep{gemmateam2024gemmaopenmodelsbased}, SmolLM-1.7B~\citep{allal2024SmolLM}, and MobiLlama-1B~\citep{thawakar2024mobillama}, which have demonstrated impressive performance while maintaining significantly smaller parameter counts. These models excel in inference speed, deployment flexibility, and privacy preservation, making them particularly suitable for resource-constrained environments.

For facilitating visual-language understanding in resource-constrained environments, researchers have developed Multi-modal SLMs by efficiently extending single-modal SLMs with visual capabilities, as exemplified by MiniCPM-V 2.0~\citep{yao2024minicpmvgpt4vlevelmllm}, Qwen2-VL~\citep{Qwen2-VL}, Phi-3-vision~\citep{abdin2024phi}, and InternVL2-2B~\citep{chen2024internvl}. These multi-modal SLMs have demonstrated remarkable capabilities in combining visual and textual understanding while maintaining the computational efficiency advantages of small-scale models, enabling their deployment in diverse applications from GUI Agent~\citep{lin2024showui} to robotics control~\citep{zheng2024tracevlavisualtraceprompting}.

While SLMs have demonstrated impressive capabilities in language understanding and multi-modal tasks, the potential of leveraging efficient models for RAG tasks remains largely unexplored. This work fills this gap by introducing a framework that enables SLMs to effectively perform RAG tasks while maintaining their inherent advantages in computational efficiency and deployment flexibility.

\textbf{Retrieval-Augmented Generation.}
Retrieval-Augmented Generation (RAG) systems enhance language models' responses by retrieving relevant knowledge from external databases~\citep{guo2024lightrag,qian2024memorag,gao2024retrievalaugmented}. The process consists of three main components: indexing, retrieval, and generation. Given a raw text set, the system first processes it into a database, then retrieves relevant information based on user query $q$, and finally generates the answer. Two primary approaches have emerged in database construction: (1) chunks-based methods~\citep{mao2020generation,qian2024memorag}, which segment texts into retrievable units, and (2) graph-based methods~\citep{guo2024lightrag,edge2024localglobalgraphrag}, which structure information as knowledge graphs.

Recent RAG implementations have evolved along these two paths. Chunks-based methods, exemplified by NaiveRAG~\citep{mao2020generation}, ChunkRAG~\citep{singh2024chunkragnovelllmchunkfiltering}, and RQ-RAG~\citep{chan2024rq}, focus on optimizing text segmentation and chunk retrieval strategies. Graph-based approaches, such as GraphRAG~\citep{edge2024localglobalgraphrag}, LightRAG~\citep{guo2024lightrag}, and SubgraphRAG~\citep{li2024simpleeffectiverolesgraphs}, leverage graph structures for better corpus comprehension and efficient retrieval. However, most existing methods require either large context windows~\citep{qian2024memorag,li2024simpleeffectiverolesgraphs} or strong semantic understanding capabilities~\citep{guo2024lightrag,edge2024localglobalgraphrag}, limiting their applicability with small and lightweight language models. This gap motivates the development of more efficient RAG systems suitable for resource-constrained scenarios.

%% file: conclusion.tex
\section{Conclusion}
We present \model, a novel RAG system designed to address the fundamental limitations of deploying small language models (SLMs) in existing retrieval-augmented generation frameworks. Through its innovative heterogeneous graph indexing and lightweight heuristic retrieval mechanisms, \model\ effectively integrates the advantages of both text-based and graph-based RAG approaches while significantly reducing the demands on language model capabilities. Our experimental results demonstrate that \model\ can achieve comparable performance to LLM-based methods even when using SLMs. Additionally, to facilitate research in this emerging field, we release the first benchmark dataset specifically designed for evaluating on-device RAG capabilities, featuring realistic personal communication scenarios and multi-constraint queries. These contributions mark an important step toward enabling private, efficient, and effective on-device RAG systems, opening new possibilities for edge device AI applications while preserving user privacy and resource efficiency.

%% file: appendix.tex
\appendix
\onecolumn

\renewcommand\thefigure{A\arabic{figure}}
\renewcommand\thetable{A\arabic{table}}
\renewcommand\theequation{A.\arabic{equation}}
\renewcommand\thetheorem{A.\arabic{theorem}}
\setcounter{table}{0}
\setcounter{figure}{0}
\setcounter{theorem}{0}
\setcounter{equation}{0}

\begin{center}
\huge {\textbf{Appendix}}    
\end{center}

\normalsize

\input{dataset_new}

%% file: dataset_new.tex
$\bullet$ \textbf{Dataset Descriptions}. The rapid growth of mobile computing has led to an unprecedented accumulation of content on personal devices, creating a pressing need for efficient on-device information retrieval and generation systems. Traditional RAG benchmarks, primarily focused on well-structured documents like Wikipedia articles or academic papers, fail to capture the unique characteristics of on-device scenarios. These scenarios present distinct challenges and characteristics:

\begin{itemize}[leftmargin=*]

\item (1) \textbf{Content Fragmentation and Context-Switching} - Unlike traditional RAG systems that process well-structured documents with clear logical flow (e.g., Wikipedia articles, academic papers), on-device RAG must handle inherently fragmented content that frequently switches between different contexts and conversation threads - a direct reflection of how people naturally communicate and interact across various digital platforms in their daily lives.

\item (2) \textbf{Temporal Nature and Evolution Patterns} - Unlike traditional RAG's static, historically complete documents, on-device RAG must handle inherently dynamic content that constantly evolves through real-time updates, ongoing conversations, and emerging social interactions. This fundamental difference in temporal dynamics creates unique challenges for on-device RAG systems, which must maintain relevance and accuracy while processing an ever-changing stream of information across various digital platforms and communication channels.

\item (3) \textbf{Digital-Physical Context Fragmentation}. Digital communication related to personal social activities and events typically presents fragmented and incomplete information, as these interactions span both online and offline contexts. Unlike traditional RAG systems that process self-contained documents, on-device content frequently captures only partial context of real-world events - text conversations might reference in-person meetings, shared experiences, or future plans without full details. This hybrid nature means digital communications often contain implicit references and assumed knowledge that only make sense with additional real-world context, requiring sophisticated context-bridging capabilities to effectively process and understand the information.

\end{itemize}

\begin{figure*}[h]
    \begin{center}
    \includegraphics[width=0.6\textwidth]{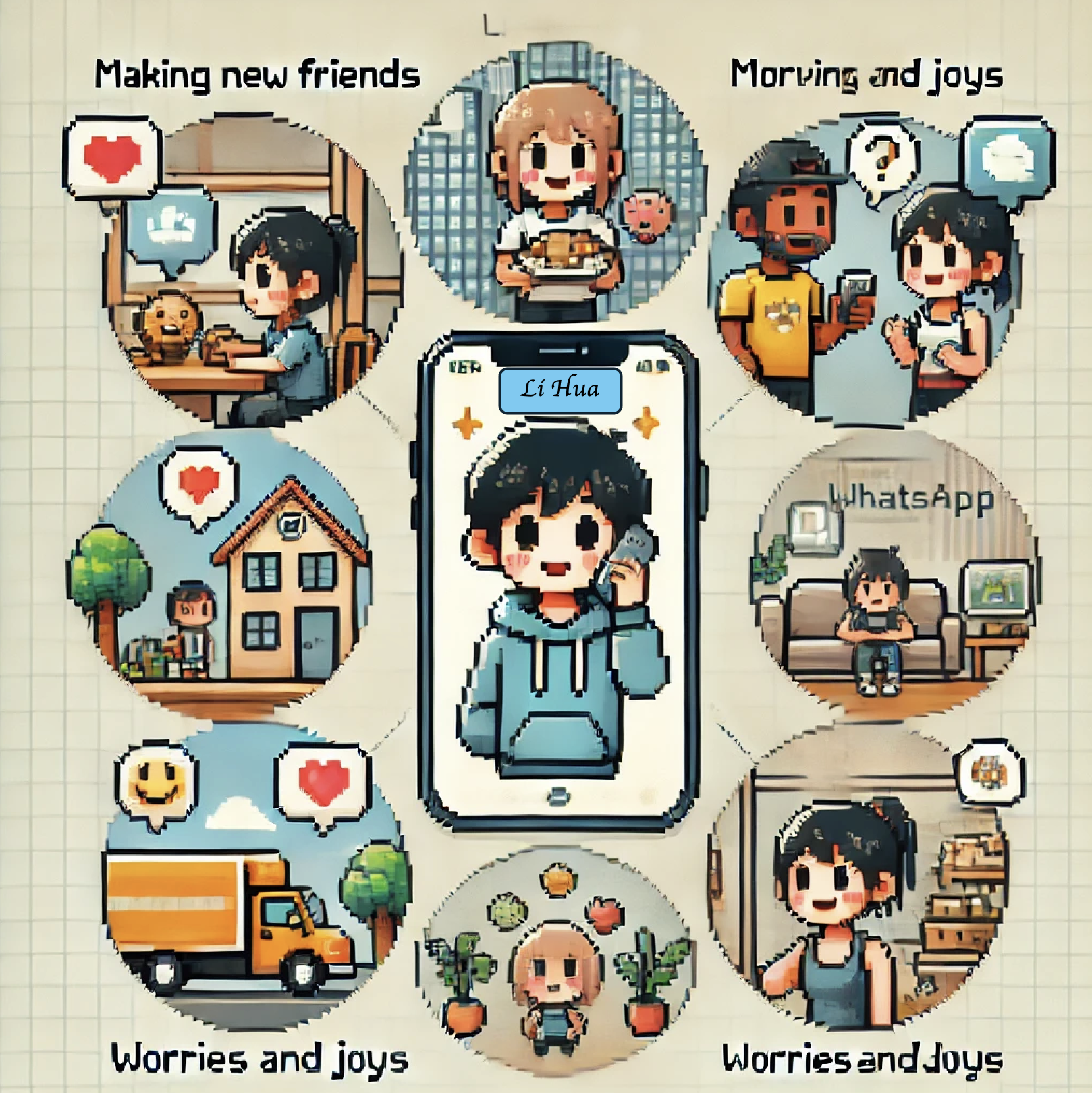}
    \end{center}
    \vspace{-1em}
    \caption{\dataset~simulates a digitally interconnected world where AI agents communicate through mobile chat applications. Through the lens of our primary subject, Li Hua, we observe and collect authentic chat interactions within this virtual social ecosystem.
    }
    \label{datasetill}
    \vspace{-0.5em}
\end{figure*}

The \dataset~dataset authentically reflects key characteristics of on-device communications, emphasizing digital-physical context fragmentation and temporal evolution patterns. Given the private nature of on-device chat data, we employed a detailed simulation approach to generate this comprehensive dataset, implementing a year-long life journey across contemporary messaging platforms. The simulation follows our protagonist, Li Hua, through both major life events and daily social interactions that naturally span between digital conversations and physical contexts.

The dataset deliberately incorporates challenging aspects typical of on-device content: conversations span multiple contexts and threads, information develops and updates over time, and messages frequently contain implicit references to offline events. These features reflect the real-world complexity of digital communications, where content evolves dynamically and often requires bridging between online and offline contexts for complete understanding. The temporal nature of interactions is carefully preserved, showing how information and relationships develop over extended periods.

The dataset includes a wide range of scenarios, from social coordination to life transitions and daily activities. For instance, weekend plans might begin with informal group chat coordination and evolve through real-time updates, while housing-related conversations span multiple threads with partial context shared across physical viewings and digital negotiations. Such scenarios demonstrate the dataset's capacity to capture both the breadth and depth of typical on-device communications.

$\bullet$ \textbf{Dataset Structure}. The \dataset~timeline begins with Li Hua's relocation to a new city, serving as a strategic starting point that naturally facilitates the simulation of expanding social networks and evolving relationships. Within this narrative framework, the conversation data is systematically organized into two primary categories: one-on-one chats and group chats.

\textbf{One-on-One Chats}\\
Li Hua, the primary agent in our dataset, engages in conversations with a diverse network of simulated friends. These conversational partners represent varied backgrounds, professions, relationships, and interests, enabling rich and authentic interactions throughout the timeline. Below is a representative one-on-one conversation extracted from \dataset~that illustrates these dynamics:

\begin{quote}
\textbf{Time: 20260819\_10:00}\\
\textbf{LiHua:} Hey Jake! It's really nice to meet you! I just love the enthusiasm you have for soccer! \\
\textbf{JakeWatson:} It's really nice to meet you, too, Li Hua! And I like your passion as well! \\
\textbf{LiHua:} Just wondering if you have some time this weekend to help me out with my dribbling skills? I could really use your expertise! \\
\textbf{JakeWatson:} Yeah, I’d love to help you out. How about Saturday afternoon? We can hit the field and work on your skills!\\
\textbf{LiHua:} Saturday afternoon sounds perfect! What time are you thinking? \\
\textbf{JakeWatson:} How about we meet around 3 PM? Should give us plenty of time to practice! Let me know if that works for you. \\
\textbf{LiHua:} 3 PM works for me! Looking forward to it. Thanks for helping out! \\
\textbf{JakeWatson:} No problem at all! It’ll be fun, and I'm excited to see you improve. Just bring some water, and we’ll have a blast! \\
\textbf{LiHua:} Definitely! I’ll be ready. See you on Saturday!
\end{quote}

\textbf{Group Chat}\\
Beyond one-on-one interactions, \dataset~captures multi-participant conversations in groups. The following excerpt from the ``TV Show Lovers'' group chat demonstrates these collective interactions:

\begin{quote}
\textbf{Time: 20260826\_18:00}\\
\textbf{EmilyBurnett:} Hey guys! Let's discuss about the main character Kendall's motivation in the TV series Succession! I think it will be fun to take a close look at him! I will go first!\\
\textbf{EmilyBurnett:} I think Kendall’s motivations are so complex! It's like he’s trying to prove himself, not just to his dad but also to the world. Plus, I feel like the pressure of the family legacy weighs heavily on him. What do you all think? \\
\textbf{Kieran:} I find it interesting how Kendall’s struggle for power ties back to his childhood. The constant rivalry and need for approval from Logan really shaped him. Do you think he can ever fully break free from that dynamic? \\
\textbf{Tamara:} Totally agree! His relationship with Logan is so toxic, but Kendall keeps coming back for more validation. It's like he’s in this endless cycle.\\
\textbf{...} \\
\textbf{LiHua:} The intensity of Kendall's journey really keeps us on edge! It's hard not to root for him despite everything, especially when you see how hurt he is. \\
\textbf{Kieran:} I really think it's a mix of both! On one hand, he craves that power and validation, but on the other, he seems desperate to carve out his own identity separate from Logan's shadow. It's such an interesting storyline, watching him fight that internal battle. \\
\textbf{...} \\
\textbf{EmilyBurnett:} Absolutely! The suspense makes it so much more thrilling. Plus, with all the character complexities, there’s never a dull moment.
\end{quote}

$\bullet$ \textbf{Event Generation with Human Oversight}.
Events serve as conversation catalysts, functioning as carefully crafted scripts that guide character interactions and dialogue topics. While GPT-4-mini occasionally provides creative inspiration, our team primarily authors these events through deliberate human curation to ensure narrative coherence and authenticity. The conversation generation process is powered by \textbf{AgentScope}~\citep{gao2024agentscope}, which transforms these event scripts into natural dialogues. Below is a representative excerpt of events from a typical week:

\begin{center}
    \renewcommand{\arraystretch}{1.5}
    \begin{tabularx}{\textwidth}{|p{2.5cm}|p{3.5cm}|X|}
    \hline
    \textbf{Time} & \textbf{Participants} & \textbf{Case} \\
    \hline
    20260818\_10:00 & Li Hua and Thane Chambers & Thane Chambers asks Li Hua which character in the game \emph{Witcher 3} Li Hua likes the best and why. \\
    \hline
    20260819\_10:00 & Li Hua and Jake Watson & Li Hua messages Jake Watson asking Jake if he has some time during the weekend to help Li Hua improve his dribbling skills. \\
    \hline
    20260820\_14:00 & Li Hua, Emily, and \newline Others in TVfan group & Emily Burnett creates a poll for the group to vote on their favorite HBO series of all time. \\
    \hline
    \end{tabularx}
\end{center}

\textbf{Query Set Design}. Our query set has two dimensions: event-based content and reasoning complexity. The event-based dimension encompasses six categories (When, Where, Who, What, How, and Yes/No questions), while the reasoning complexity distinguishes between single-hop and multi-hop queries based on required inferential steps. The following examples illustrate these diverse query types:

\begin{center}
    \renewcommand{\arraystretch}{1.5}

    \begin{tabularx}{\textwidth}{|p{1cm}|p{3.8cm}|p{2.8cm}|p{2.4cm}|X|}
    \hline
    \textbf{NO.} & \textbf{Question} & \textbf{Gold} \newline \textbf{Answer} & \textbf{Support} \newline \textbf{ Documents} & \textbf{Type} \\
    \hline
    \textbf{1} & What hobbies does Li Hua have? & photography, guitar, fitness, video game, TV show, soccer & NA & What \\
    \hline
    \textbf{2} & What is the Wi-Fi password at Li Hua's house? & Family123 & 20260106\_09:00 & What\\
    \hline
    \textbf{3} & Who is Li Hua sending a fruit basket to? & Adam & 20261027\_17:00 & Who\\
    \hline
    \textbf{4} & Did Wolfgang ask Li Hua about watching "Star Wars: A New Hope" after he asked Li Hua about going to see "Overwatch 3"? & Yes & 20260121\_13:00 \textless and\textgreater \newline 20261009\_17:00 & YesNo\\
    \hline
    \end{tabularx}
\end{center}